\documentclass[lettersize,journal]{IEEEtran}
\usepackage{amsmath,amsfonts}
\usepackage{algorithmic}
\usepackage{algorithm}
\usepackage{array}
\usepackage[caption=false,font=normalsize,labelfont=sf,textfont=sf]{subfig}
\usepackage{textcomp}
\usepackage{stfloats}
\usepackage{url}
\usepackage{verbatim}
\usepackage{graphicx}
\usepackage{cite}
\usepackage{xcolor}
\usepackage{bbding}
\usepackage{caption}
\usepackage{subcaption}
\usepackage{multirow}
\usepackage[normalem]{ulem}
\begin{document}

\title{LC-TTFS: Towards Lossless Network Conversion for Spiking Neural Networks with TTFS Coding}

\author{Qu~Yang{*}, Malu~Zhang{*}, Jibin~Wu,
	Kay~Chen~Tan,~\IEEEmembership{Fellow,~IEEE}
	and~Haizhou~Li,~\IEEEmembership{Fellow,~IEEE,}
    \IEEEcompsocitemizethanks{This research work was supported in part by the IAF, A*STAR, SOITEC, NXP and National University of Singapore under FD-fAbrICS: Joint Lab for FD-SOI Always-on Intelligent Connected Systems (Award I2001E0053), the Agency for Science, Technology and Research (A*STAR) under its AME Programmatic Funding Scheme (Project No. A18A2b0046), A*STAR under its RIE 2020 Advanced Manufacturing and Engineering Human (AME) Programmatic Grant (Grant No. A1687b0033), and by the Hong Kong Polytechnic University under Project P0043563, P0046094, and P0046810. (* Contributed equally in this work, Corresponding Author: J.~Wu).
    
	\IEEEcompsocthanksitem Q.~Yang is with the Department of Electrical and Computer Engineering, National University of Singapore, Singapore. 
     \IEEEcompsocthanksitem H. Li is with Shenzhen Research Institute of Big Data, School of Data Science, The Chinese University of Hong Kong, Shenzhen (CUHK-Shenzhen), China. 
    \IEEEcompsocthanksitem M.~Zhang is with University of Electronic Science and Technology of China, China.
	\IEEEcompsocthanksitem J.~Wu and K.~C.~Tan are with the Department of Computing, The Hong Kong Polytechnic University, Hong Kong SAR, China. }
}



\maketitle

\begin{abstract}
The biological neurons use precise spike times, in addition to the spike firing rate, to communicate with each other. The time-to-first-spike (TTFS) coding is inspired by such biological observation. However, there is a lack of effective solutions for training TTFS-based spiking neural network (SNN). In this paper, we put forward a simple yet effective network conversion algorithm, which is referred to as LC-TTFS, by addressing two main problems that hinder an effective conversion from a high-performance artificial neural network (ANN) to a TTFS-based SNN. We show that our algorithm can achieve a near-perfect mapping between the activation values of an ANN and the spike times of an SNN on a number of challenging AI tasks, including image classification, image reconstruction, and speech enhancement. With TTFS coding, \textcolor{black}{we can achieve up to orders of magnitude saving in computation over ANN and other rate-based SNNs.} The study, therefore, paves the way for deploying ultra-low-power TTFS-based SNNs on power-constrained edge computing platforms.
\end{abstract}

\begin{IEEEkeywords}
Deep Spiking Neural Network, Time-to-first-spike Coding, ANN-to-SNN Conversion, Image Classification, Image Reconstruction, Speech Enhancement
\end{IEEEkeywords}

\section{Introduction}
\label{introduction}
\IEEEPARstart{O}{ver} the last decade, we have witnessed tremendous progress in artificial intelligence technologies, \textcolor{black}{that include computer vision \cite{krizhevsky2012imagenet, he2016deep, 8981937, 10244119}, speech processing \cite{hannun2014deep, oord2016wavenet}, natural language processing \cite{hirschberg2015advances, devlin2018bert}, and robotics \cite{silver2017mastering, 9486868}.} However, the core computational model behind this revolution, i.e., artificial neural network (ANN), is computationally expensive to operate. This prompts the researchers to improve the computational efficiency of ANNs, for example, through model compression \cite{hinton2015distilling, hubara2017quantized}, network accelerator~\cite{aimar2018nullhop}, and reduction of on-chip data movements \cite{chen2016eyeriss}. Nevertheless, the high computational cost remains a major roadblock to the deployment of ANNs on power-constrained platforms, such as wearable and mobile devices \cite{tan2019efficientnet}.

The human brains evolve over many millennia under strong ecological pressure to be highly efficient and effective, therefore, it is worthwhile to look into the computation principles adopted by biological neural networks. Motivated by this, the spiking neural networks (SNNs), which were initially introduced to simulate neural computations \cite{maass1997networks}, are now considered a power-efficient alternative to the mainstream ANNs, with a great potential to become the solution for power-constrained platforms. 

\textcolor{black}{SNNs, which emulate the information processing mechanism of biological neural networks \cite{hodgkin1952currents, dayan2001theoretical, izhikevich2003simple, brette2005adaptive, 9717282}, represent and transmit information through asynchronous action potentials or spikes.} \textcolor{black}{Due to the complex spatial-temporal dependency of spike trains and the discontinuity at the spike generation time, the canonical back-propagation (BP) algorithm is not directly applicable to the training of SNNs \cite{pfeiffer2018deep, 9274500}}. The surrogate gradient learning method \cite{neftci2019surrogate} has been introduced recently to address these problems. It models the spiking neuron as a self-recurrent neural network that explicitly captures the spatial-temporal dependency between input and output spike trains. Furthermore, the continuous surrogate gradient functions are introduced during gradient back-propagation that effectively addresses the discontinuity issue at the spike generation time. \textcolor{black}{Despite much progress on a host of machine learning and neuromorphic benchmarks \cite{bellec2018long, wu2019direct,shrestha2018slayer,wu2019hybrid,fang2021deep,duan2022temporal, 9406125}, it is computationally prohibitive to exactly model the temporal dynamics of SNNs due to exorbitant memory requirement, even for a network of less than ten layers \cite{shrestha2018slayer}.} Besides, this method suffers from the vanishing and exploding gradient problem that is notorious for recurrent neural networks, making long-range temporal credit assignments ineffective. A more biologically plausible way entails considering propagating spikes only when the neuron fires spikes, which reduces the overall number of gradient propagations on neuromorphic hardware~\cite{zhu2022training}. Zhu et al.~\cite{zhu2022training} have rigorously proved that event-based propagation allocates gradients from one output spike to its corresponding input spikes, thus preserving the sum of gradients between layers. Armed with this insight, they successfully trained SNNs with temporal gradients on the CIFAR-100 dataset for the first time.

In another vein of research, ANN-to-SNN conversion methods are introduced as an alternative solution to address the difficulties in direct SNN training. A large body of these network conversion methods takes the firing rate of spiking neurons to approximate the activation value of analog neurons used in the ANN, which we refer to as \textit{rate conversion} in this paper. By carefully determining the firing threshold of spiking neurons or the weight normalization factor, the pre-trained network parameters of ANN can be directly transferred to the SNN with an equivalent network structure \cite{wu2019deep, esser2015backpropagation, hunsberger2015spiking, esser2016convolutional, o2013real, liu2017noisy, diehl2015fast, diehl2016truehappiness, rueckauer2017conversion, rueckauer2018conversion, han2020rmp,bu2022optimal,hao2023reducing,bu2022optimized,ding2021optimal}. In this way, we can avoid the expensive spatio-temporal credit assignments in surrogate gradient learning. The rate conversion methods map ANN to SNN with high precision on major machine learning benchmarks, such as ImageNet-12 \cite{sengupta2019going, han2020rmp, rathi2020enabling, zheng2020going, yan2021near}, with a high latency~\cite{rueckauer2018conversion} due to the requirement of a large simulation time window. The benefits that rate-based SNNs bring are hence limited. 

\begin{figure*}[t]
    \centering  
	\subfloat[]{
	\includegraphics[scale=0.40]{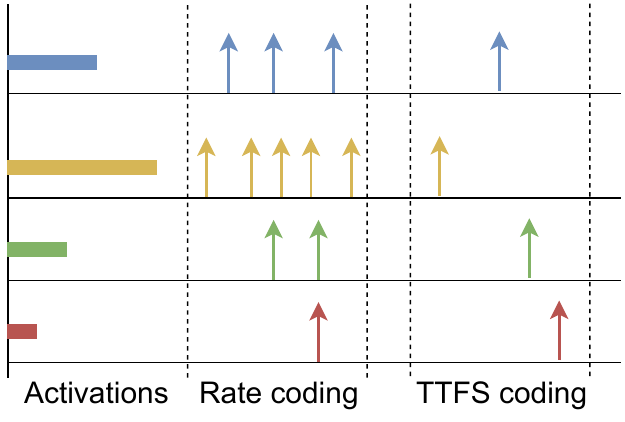}}\hspace{0em}
	\subfloat[]{
	\includegraphics[scale=0.6]{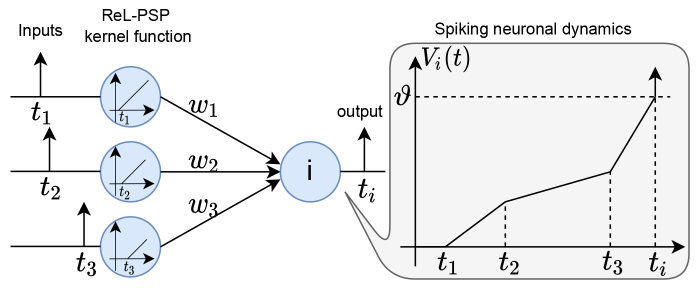}}\hspace{0em}
	\subfloat[]{
	\includegraphics[scale=0.55]{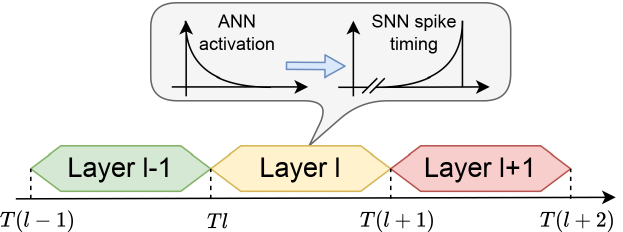}}\hspace{0em}
    
    \caption{(a) Comparison of rate and TTFS coding schemes. (b) Illustration of spiking neuronal dynamics with ReL-PSP kernel function. (c) Illustration of inference pipeline of the TTFS converted SNN, wherein each layer operates in consecutive but non-overlapping time windows. The inset on top shows the data distribution of activation values in an ANN and spike times of the converted SNN. The activation values are mapped to spike times in a one-to-one correspondence following the TTFS encoding scheme.}
    \label{fig: background_plots}
\end{figure*}

It has long been identified in the biological neural systems that the precise spike firing time, in addition to the spike firing rate, carries a significant amount of information \cite{gutig2006tempotron}. Based on these insights, the time-to-first-spike (TTFS) coding scheme was formulated~\cite{thorpe2001spike,kheradpisheh2020temporal,kheradpisheh2022bs4nn,mirsadeghi2022ds4nn}, where only one single spike is allowed within each time window as shown in Fig. \ref{fig: background_plots} (a), and a stronger stimulus is transduced into an earlier firing spike. In this way, with a fewer number of spikes, the TTFS coding scheme is expected to be more efficient computationally than rate-based coding.

Embracing the TTFS coding, Rueckauer et al. \cite{rueckauer2018conversion} proposed an ANN-to-SNN conversion algorithm, where the activation value of ANN was treated as the instantaneous firing rate, and subsequently converted to the equivalent spike time in SNN. However, the scalability of this conversion algorithm to deeper network architecture was not demonstrated. The TDSNN \cite{zhang2019tdsnn} algorithm has been proposed to achieve better scalability, which introduces an auxiliary ticking neuron for each operating spiking neuron and a TTFS-like reverse coding scheme. However, the additional spikes generated from auxiliary ticking neurons adversely affect the model efficiency. \textcolor{black}{In contrast, our method uses the same network architecture as the ANN, which does not require any additional neurons or spikes.}
Recently, the T2FSNN \cite{park2020t2fsnn} algorithm has been introduced with improved conversion performance on TTFS-based SNNs, while their results still lag behind rate-based SNNs. \textcolor{black}{Moreover, their kernel-based dynamic threshold is considered more computationally expensive on hardware implementation than ours layer-wise dynamic thresholds.}
Similar to the TDSNN work, Han and Roy \cite{han2020deep} proposed a Temporal-Switch-Coding (TSC) scheme and a corresponding TSC spiking neuron model. However, this coding scheme requires two spikes to encode the information of each analog neuron, whereby the information is represented as the time difference between these two spikes. Besides, the introduced TSC neuron model is computationally more expensive than the Integrate-and-Fire neuron model adopted in other works. In what follows, in contrast to the rate conversion introduced earlier, we refer to the ANN-to-SNN conversion based on the TTFS coding as \textit{TTFS conversion}.

In this work, we aim to bridge the accuracy gap between TTFS conversion and rate conversion, thereby fully realizing the computational advantages of SNNs on power-constrained platforms. Toward this goal, we make the following three contributions: 
\begin{enumerate}
    \item We perform a comprehensive analysis on the problems underlying TTFS conversion, \textcolor{black}{including temporal dynamics problem and time-warping problem.}
    \item We propose a simple yet effective TTFS conversion algorithm that can effectively address all the above problems. As shown in Fig. \ref{fig: background_plots} (c), it establishes a near-perfect mapping between activation values of ANNs and spike times of SNNs, which leads to a near-lossless TTFS conversion. 
    \item We successfully implement the TTFS conversion algorithm for image classification, image reconstruction, and speech enhancement tasks. To the best of our knowledge, this is the first work that applies TTFS-based SNN in solving challenging signal reconstruction tasks.
\end{enumerate}

The rest of this paper is organized as follows: In Section \ref{problem}, we introduce the research problems to set the stage for our study. In Section \ref{method}, we present the proposed TTFS conversion algorithm to address the identified problems. In Section \ref{exp}, we first evaluate the proposed conversion algorithm on the image classification task. Then, we thoroughly evaluate the effectiveness of the proposed algorithm through a series of ablation studies. In Section \ref{regression}, we further evaluate the proposed algorithm on two signal reconstruction tasks, i.e. image reconstruction and speech enhancement. Finally, Section \ref{conclusion} concludes the study.

\section{Preliminaries and Problem Analysis}
\label{problem}
We begin by introducing the analog and spiking neuron models, as well as the TTFS coding scheme. We then analyze the problems underlying TTFS conversion.

\subsection{Preliminaries}
\subsubsection{Analog neuron model}
\label{analog}
In ANN-to-SNN conversion, an ANN is first trained, wherein analog neurons are employed and formulated as follows,
\begin{equation}
a^l_i=f(\sum_i^Nw_{ij}^{l}a_j^{l-1} + b_i^l)
\label{analog_neuron}
\end{equation}
where $a_j^{l-1}$ and $a^l_i$ are the input and output of neuron $i$ at layer $l$, respectively. $w_{ij}^{l}$ is the synaptic weight between pre-synaptic neuron $j$ and post-synaptic neuron $i$, and $b_i^l$ is the bias term of neuron $i$. $f(\cdot)$ denotes the activation function, for which we use a modified version of the Rectified Linear Unit (ReLU) function. Specifically, we clamp the activation value to be within $[0,1]$, similar to the ReLU6 function \cite{krizhevsky2010convolutional}, and we refer to it as ReLU1 hereafter. As will be explained in the following section, this ensures a one-to-one correspondence can be established between ANN and SNN for lossless TTFS conversion.

\subsubsection{Spiking neuron model}
\label{spiking}
For SNN, to be converted from the pre-trained ANN, we employ the Rectified Linear Postsynaptic Potential (ReL-PSP) spiking neuron model proposed in \cite{Zhang2020}, whose membrane potential $V_i^l(t)$ can be expressed as
\begin{equation}
V^l_i(t)=\sum_{j=1}^{N^{l-1}} w_{ij}^{l}K(t-t_j^{l-1})
\label{neuronmodel}
\end{equation}
where $K(\cdot)$ refers to the PSP kernel function, which is defined as 
\begin{equation}K(t-t_j^{l-1}) =
\begin{cases}
t-t_j^{l-1}\quad \text{if} \quad t>t_j^{l-1}\\ 
0 \quad\quad\quad\quad\text{otherwise}
\end{cases}
\label{eq: ReL-PSP}
\end{equation}

\noindent For $t > t_j^{l-1}$, Eq. (\ref{neuronmodel}) can be further simplified as
\begin{equation}
\label{V_t}
V^l_i(t)=\sum_{j=1}^{N^{l-1}} w_{ij}^{l} (t-t_j^{l-1})
\end{equation}

\noindent The neuron $i$ fires a spike once its membrane potential exceeds the firing threshold $\vartheta$, whose spike time $t_i^l$ is defined as
\begin{equation}
    t_i^l = \mathcal{F} \left\{ t |V_i^l(t)=\vartheta, t \geq 0 \right\}
    \label{spikegenerate}
\end{equation}
The neuronal dynamics of the ReL-PSP spiking neuron model are illustrated in Fig. \ref{fig: background_plots} (b).
Without loss of generality, we set the firing threshold $\vartheta$ to 1 in this work.

\subsubsection{TTFS encoding scheme}
\label{TTFS_encoding}
To encode the first spike time of a ReL-PSP spiking neuron with the activation value of a ReLU1 neuron, we follow the TTFS encoding scheme. For each layer \textcolor{black}{that with separate time window}, as shown in Fig. \ref{fig: background_plots} (a) and \textcolor{black}{(c)}, we encode the activation value $a_i^l$ into the spike time $t_i^l$ as per
\begin{equation}
\label{eq:encoding}
\frac{t_i^l-t_{min}^l}{t_{max}^l-t_{min}^l} = 1 - a_i^l
\end{equation}
\noindent where $t_{max}^l$ and $t_{min}^l$ are the maximum and minimum permissible spike time of SNN layer $l$. We define the time window $T^l = t_{max}^l - t_{min}^l$. Without loss of generalizability, we fix this value to be 1 for all layers, i.e., $T = 1$. Hence, we can establish the following encoding function 
\begin{equation}
t_i^{l}= l+1 - a_{i,n}^{l},
\label{eq:latencycoding}
\end{equation}

\subsection{TTFS Conversion Problems}
\subsubsection{Temporal dynamics problem}
\label{sec:premature_spike_problem}
\label{premature}

Following the above TTFS encoding scheme, a larger activation value in the ANN layer shall be encoded into an earlier spike in the corresponding SNN layer, and vice versa. The additional temporal dynamics introduced during the conversion process may, however, give rise to missing spikes and premature spikes. The missing spike problem happens, during ANN-to-SNN conversion, when an analog neuron receives multiple tiny weighted inputs, while their corresponding spikes as a whole are insufficient to trigger an output spike within the given time window. In contrast, the premature spike problem\textcolor{black}{, which has been also described in \cite{rueckauer2018conversion},} happens when the positive weighted spikes arrive earlier than the negative ones, causing the actual number of output spikes more than expected.

\begin{figure}[t]
    \centering  
	\subfloat[ANN]{
	\includegraphics[scale=0.6]{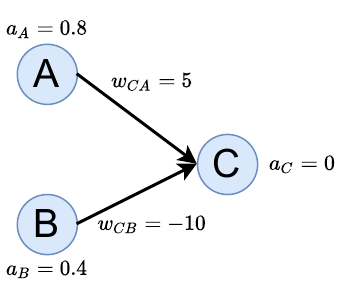}}\hspace{0em}
	\subfloat[SNN]{
	\includegraphics[scale=0.6]{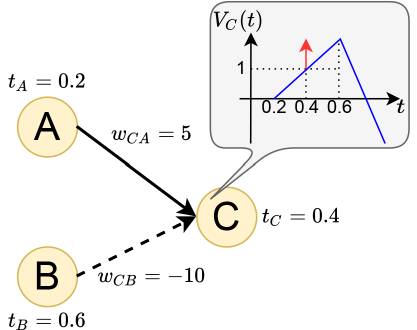}}\hspace{0em}
    
    \caption{Illustration of the premature spike problem. (a) For ANN, the net inputs from neurons A and B are summed to $0$, causing neuron C to remain inactivated. (b) For SNN, after TTFS conversion, the earlier spike from neuron A will cause neuron C to fire a `premature' spike at $t_c = 0.4$, before the arrival of the inhibition input from neuron B.}
    \label{fig:premature}
\end{figure}

To better understand the premature spike problem, let's consider a network formed by two input neurons $A$ and $B$ that is connected to an output neuron $C$, with synaptic weights $w_{CA} = 5$ and $w_{CB} = -10$. As shown in Fig. \ref{fig:premature}(a), assuming the activation value of analog neurons $A$ and $B$ is $a_A = 0.8$ and $a_B = 0.4$. As these two weighted inputs cancel out each other, the net input received by the neuron $C$ will be $0$. According to Eq. (\ref{eq:encoding}), the converted spike time of spiking neurons $A$ and $B$ are $t_A = 0.2$ and $t_B = 0.6$. Assuming the firing threshold is $1$, the earlier input spike from neuron $A$ will trigger neuron $C$ to fire a spike at $t_C = 0.4$, before receiving the inhibition input from neuron $B$. In this example, both neurons $A$ and $B$ contribute to neuron $C$ in the ANN, while the contribution of neuron B has been discarded in the SNN due to the additional temporal dynamics introduced after the TTFS conversion.

We refer to both missing spike and premature spike problems as the temporal dynamics problem, which will compound across network layers and may eventually lead to a fatal mismatch between the outputs of ANN and SNN. To eliminate this problem, we propose a dynamic firing threshold mechanism to ensure spiking neurons in the $l$-th layer only fire within the permissible time window $[lT, (l+1)T)$, after all their input spikes are received. This ensures the contributions from pre-synaptic spiking neurons are fully considered by the post-synaptic spiking neuron. The details of this mechanism will be explained in Section \ref{dynamicTH}.

\subsubsection{Time-warping problem}
\label{mismatch}
As we already introduced in Section \ref{spiking}, the spiking neuron will fire a spike when the membrane potential $V_i^l(t)$ reaches the firing threshold $\vartheta$. According to Eq. (\ref{spikegenerate}), the spike time $t_i^l$ can be calculated as
\begin{equation}
    t_i^l=\frac{\vartheta+\sum_{j=0}^{N^{l-1}}w_{ij}^lt_j^{l-1}}{\sum_{j=0}^{N^{l-1}}w_{ij}^l}
    \label{spikevsweight}
\end{equation}

\noindent Following Eq. (\ref{eq:encoding}), the activation value $a_j^{l-1}$ of the analog neuron should be encoded into spike time $t_j^{l-1}$ as per 
\begin{equation}
    t_j^{l-1}= l - a_{j}^{l-1}
    \label{latencycoding}
\end{equation} 
Taking Eq. (\ref{latencycoding}) into Eq. (\ref{spikevsweight}), we can establish the following relationship between the activation value $a_j^{l-1}$ and the spike time $t_i^l$
\begin{equation}
\label{eq:tj_ai}
t_i^l = \frac{\vartheta + l \times \sum_{j=0}^{N^{l-1}}{w_{ij}^{l}} - \sum_{j=0}^{N^{l-1}}{w_{ij}^{l} a_j^{l-1}}}{\sum_{j=0}^{N^{l-1}}{w_{ij}^{l}}}
\end{equation}

\begin{figure*}[t]
    \centering  
	\subfloat[]{
	\includegraphics[scale=0.85]{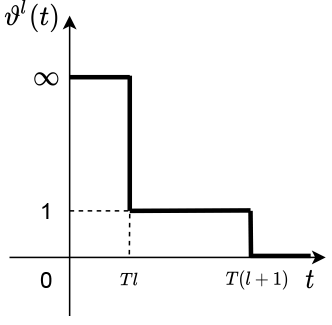}}\hspace{0em}
	\subfloat[]{
	\includegraphics[scale=0.85]{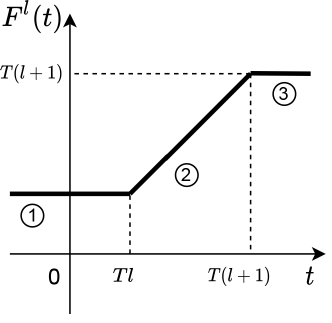}}\hspace{0em}
	\subfloat[]{
	\includegraphics[scale=0.85]{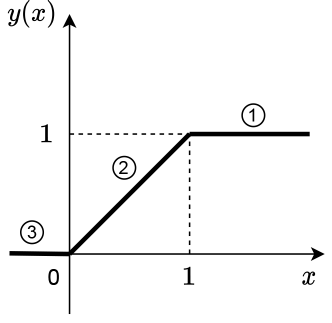}}\hspace{0em}
    
    \caption{(a) Illustration of the dynamic firing threshold. (b) Illustration of the effective spike time transformation achieved by the dynamic firing threshold. The x- and y-axis represent the spike time before and after applying the dynamic firing threshold, respectively. (c) Illustration of the ReLU1 activation function used in the ANN. Note that the activation values of analog neurons in (c) can be mapped to spike times in (b) in a one-to-one correspondence, while the order of the mapping is reversed following the TTFS encoding scheme.}
    \label{fig: combine_plots}
\end{figure*}

\noindent According to Eq. (\ref{eq:tj_ai}), the spike time $t_i^l$ depends on both weighted inputs $\sum_{j=0}^{N^{l-1}}{w_{ij}^{l} a_j^{l-1}}$ and weight sum $\sum_{j=0}^{N^{l-1}}{w_{ij}^{l}}$. When transferring an activation value $a_j^{l-1}$  from an analog neuron to a spiking neuron, this additional dependency on the weight sum will cause a significant mismatch between the ANN and SNN outputs if not properly addressed. 
\textcolor{black}{Ideally, identical inputs should produce consistent outputs in $t_i^l$. Yet, variations in the weight sum for each neuron $i$ cause deviations in the output for the same input value.}
We refer to this problem as the \textit{time-warping problem}. As will be introduced in Section \ref{hardCons}, we propose a weight regularization strategy to ensure the weight sum equals $1$, thereby eliminating this time-warping problem.

\section{LC-TTFS conversion algorithm}
\label{method}
\subsection{Solve temporal dynamics problem with dynamic threshold}
\label{dynamicTH}
As discussed in Section \ref{sec:premature_spike_problem}, the temporal dynamics problem will result in a mismatch between the ANN and SNN outputs. To address this problem, we propose a dynamic firing threshold for spiking neurons. As shown in Fig. \ref{fig: combine_plots}(a), the dynamic firing threshold $\vartheta^l(t)$ for neurons in the $l$-th layer is determined according to the following piecewise linear function
\begin{equation}\vartheta^l(t) =
\begin{cases}
\infty \quad \hspace{0.18cm}\text{if} \quad  t < T l\\ 
1 \quad \quad \text{if} \quad t\in[Tl,T(l+1))\\ 
\textcolor{black}{-\infty} \hspace{0.25cm} \text{if} \quad t \geq T (l+1)\\ 
\end{cases}
\label{eq:dyn_th}
\end{equation}
\noindent

The role of the proposed dynamic firing threshold is to set the spike time outside the permissible time window to the two boundary values. This is equivalent to applying a transformation function $F^l(t)$, defined as in Eq. (\ref{eq:snn_trans}) and illustrated in Fig. \ref{fig: combine_plots}(b). Following this dynamic firing threshold, the earliest spike time for neurons in the $l$-th layer is $Tl$, and the latest spike should fire before $T(l+1)$. As such, the time window is non-overlapping for each layer, and spiking neurons at layer $l$ only start to fire after all input spikes from layer $l-1$ are being integrated, therefore, overcoming the temporal dynamics problem.
\begin{equation}F^l(t) =
\begin{cases}
Tl \quad \hspace{0.85cm} \text{if} \quad t < Tl\\ 
t \quad \hspace{1.10cm} \text{if} \quad t\in[Tl,T(l+1))\\
T(l+1) \quad \text{if} \quad t \geq T(l+1)\\ 
\end{cases}
\label{eq:snn_trans}
\end{equation}

\textcolor{black}{It is worth noting that there are two main differences between our proposed dynamic threshold and the method proposed in \cite{rueckauer2018conversion}. Concretely, 1) the dynamic threshold introduced in \cite{rueckauer2018conversion} depends on the weight of each neuron, resulting in the firing threshold varying from neuron to neuron that will cause significant hardware overhead. In contrast, our method shares one firing threshold for all neurons in the same layer. 2). our method does not require calculating the missing spikes which is computationally expensive.}

The dynamic firing threshold ensures that the spiking neurons fire only within their permissible time window. To achieve a lossless conversion from the ANN, the activation value of each ANN layer should also be bounded within a particular interval. To this end, we employ the ReLU1 activation function, which is formulated as follows
\begin{equation}y(x) =
\begin{cases}
0 \quad \hspace{0.12cm} \text{if} \quad x \leq 0 \\ 
x \quad \hspace{0.11cm}\text{if} \quad x\in (0,1]\\
1 \quad \hspace{0.12cm}\text{if} \quad x > 1\\ 
\end{cases}
\label{ReL-PSP}
\end{equation}

As shown in Figs. \ref{fig: combine_plots}(b) and \ref{fig: combine_plots}(c), with the proposed ReLU1 function, the ANN activation region is one-to-one mapped to the spike time region. The ablation study performed in Section \ref{ab_dynamic} highlights the necessity of this dynamic threshold mechanism toward a lossless TTFS conversion.

\subsection{Solve time-warping problem with weight regularization}
\label{hardCons}

\begin{figure}[t]
    \centering  
	\subfloat{
	\includegraphics[scale=0.35]{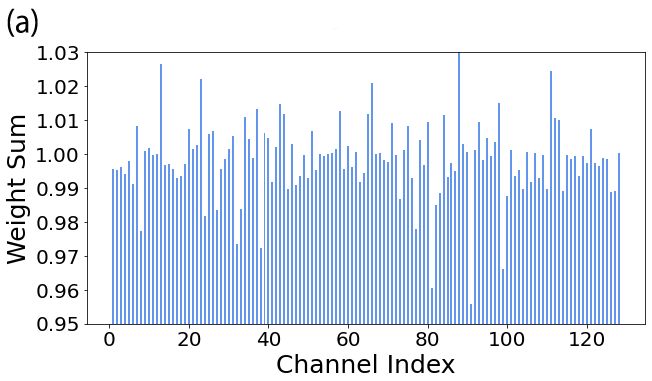}}\hspace{0em}
	
	\subfloat{
	\includegraphics[scale=0.35]{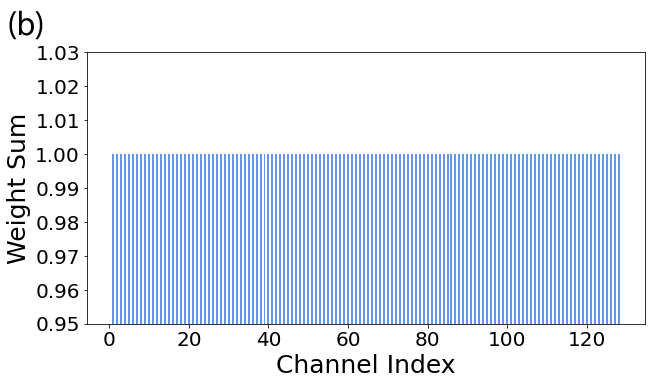}}\hspace{0em}
    \caption{Illustration of weight sum distributions of a randomly selected network layer, trained on the CIFAR-10 dataset, after imposing the (a) soft constraint, and (b) both soft and hard constraints.}
    \label{fig:soft n hard}
\end{figure}

To deal with the time-warping problem introduced in Section \ref{mismatch}, we proposed a two-step weight regularization strategy to ensure the weight sum in Eq. (\ref{eq:tj_ai}) will take a constant value of 1. In the first step, we impose a soft constraint by penalizing those weight sums that are not equal to 1 with an L1 loss function $\mathcal{L}_{W}$ that is defined as follows
\begin{equation}
\label{eq:w_sum_loss}
\mathcal{L}_{W} = \sum_{l=0}^{L-1}\sum_{i=0}^{N^l}\left|{\sum_{j=0}^{N^{l-1}}{w_{ij}^{l}}}- 1\right|
\end{equation}
As shown in Fig. \ref{fig:soft n hard}(a), this step drives the overall weight sum distribution towards 1, which can largely alleviate the time-warping problem. However, these seemingly small deviations from the ideal value of 1 will compound across layers and significantly deteriorate the performance of converted SNNs. 

To fully resolve the time-warping problem, we further impose a hard constraint by distributing the weight sum deviation evenly across all contributing synapses for each neuron. As shown in Fig. \ref{fig:soft n hard}(b), by introducing soft and hard constraints during ANN pre-training, we ensure all weight sums are exactly equal to 1. As the soft constraint already drives the weight sum to approach the ideal value, the additional hard constraint has little interference to the learning dynamics and network convergence. This has been confirmed by the ablation study that will be introduced in Section \ref{ablation}. 

\textcolor{black}{
With the time-warping problem resolved, we now can establish the relationship between $t_i^l$ and $a_i^l$. Considering weight regularization, we obtain
\begin{equation}
\label{eq:til}
	t_i^l = 1 + l - \sum_{j=0}^{N^{l-1}}w_{ij}^la_j^{l-1},   
\end{equation}
here, $\sum_{j=0}^{N^{l-1}}{w_{ij}^{l}} = 1$ and  $\vartheta = 1$. 
With the Eqns. (\ref{eq:snn_trans}, \ref{ReL-PSP}) and visualization in Fig.\ref{fig: combine_plots} (b,c), we yield
\begin{equation}
\label{eq:Fx}
	F^l(x) = T (l+1) - y(T (l+1)-x). 
\end{equation}
As mentioned in Section \ref{TTFS_encoding}, we adopt $T=1$ in this work, thereby obtaining 
\begin{equation}
\label{eq:Fx_}
	F^l(x) = l+1 - y(l+1-x).  
\end{equation}
Substitute Eqn. (\ref{eq:til}) into Eqn. (\ref{eq:Fx_}), we have 
\begin{equation}
\begin{split}    
	F^l(t_i^l) &= l+1 - y(l+1-t_i^l) \\
                   &= l+1 - y(l+1-1 - l + \sum_{j=0}^{N^{l-1}}w_{ij}^la_j^{l-1}) \\
                   &= l+1 - a_i^l
\end{split}
\end{equation}
Since $F^l(t_i^l) = t_i^l$ holds true within the permissible time window, we ultimately obtain
\begin{equation}
	t_i^l = l+1 - a_i^l,
\end{equation}
}

\textcolor{black}{
This result is consistent with the encoding scheme for neurons in layer $l-1$, as depicted by Eqn. (9). However, it introduces a T steps shift to have a non-overlapping window.}

\subsection{Pre-activation normalization strategy}
\label{pre_acti}
The BN is an important technique to accelerate the deep neural network training, which normalizes the pre-activation of each layer to follow normal distribution so as to mitigate the internal covariate shift problem.  
\textcolor{black}{
We indeed concur that the bias term from an ANN can be transposed into an SNN by injecting a constant input current at the beginning of each applicable time window. This strategy has proven effective in addressing the internal covariate shift issue in many cases \cite{wu2021progressive, yan2021near}.
However, in this work, this approach is not directly applicable to our proposed TTFS conversion method. The reason for this lies in the fact that once the BN parameters are merged with the weight sum constraint inherent in the pre-trained ANN model, the constraint becomes invalidated. Essentially, the preservation of this constraint is crucial for the functionality of our proposed framework.
}

To compensate for the absence of BN, we propose to normalize the pre-activation distribution of each layer implicitly by introducing a new loss term $\mathcal{L}_{A}$ as given in Eq. (\ref{eq:pre_acti_loss}). Since we expect the activation values to lie within $(0, 1]$ as desired for ReLU1, therefore, we apply an L1 loss to the pre-activation of each layer, encouraging them to fit a normal distribution with zero mean and standard deviation of 1/3, i.e., $\mathcal{N}(0,1/9)$. This ensures $99.7\%$ pre-activation values will lie within the interval $[-1, 1]$, such that the activation values will mostly lie within $(0, 1]$. 
\begin{equation}
\label{eq:pre_acti_loss}
\mathcal{L}_{A} = \sum_{l=0}^{L-1}\sum_{i=0}^{N^l}\left|{\sum_{j=0}^{N^{l-1}}w_{ij}^{l}a_j^{l-1}} - A_{i}^l \right|
\end{equation}
where $A^{l}_i$ is a vector, wherein entry $i$ is the normalized form, draw from $\mathcal{N}(0,1/9)$, of the pre-activation $\sum_{j=0}^{N^{l-1}}w_{ij}^{l}a_j^{l-1}$.

\begin{figure}[t]
\centering
\includegraphics[scale=0.35]{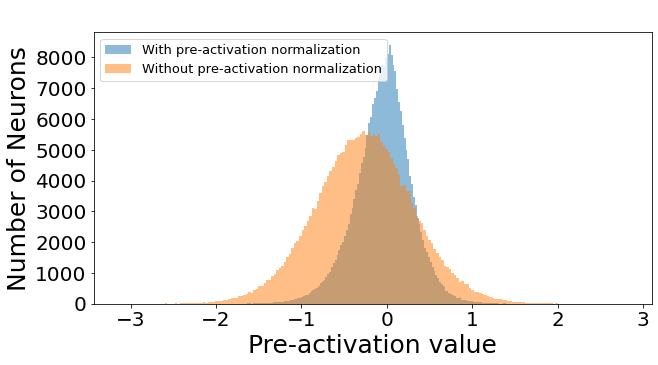}
\caption{Comparison of the pre-activation value distributions with (blue) and without (brown) applying the pre-activation normalization technique proposed in this work. Note that the data distribution approaches a normal distribution of N(0,1/9) after applying the pre-activation normalization. The data is extracted from a randomly selected network layer trained on the CIFAR-10 dataset. It is better to view this figure in color.}
\label{fig: pre_acti}
\end{figure}

In Fig. \ref{fig: pre_acti}, we show an example of how the pre-activation normalization strategy effectively drives the distribution of pre-activation values towards a normal distribution of $\mathcal{N}(0,1/9)$. Without the normalization, the pre-activation values are skewed towards a mean of $-0.3$. The effectiveness of this strategy in pre-training high-performance ANNs will be further demonstrated in our ablation study introduced in Section \ref{ab_pre-acti}.

\subsection{Overall LC-TTFS algorithm}
The proposed LC-TTFS conversion algorithm consists of two stages. In the first stage, we pre-train an ANN with the constraints described above, and the overall loss function is defined as follows
\begin{equation}
\label{eq:overall_loss}
\mathcal{L} = \mathcal{L}_{ce} + \lambda_W\mathcal{L}_{W} + \lambda_A\mathcal{L}_{A}
\end{equation}
where the $\mathcal{L}_{ce}$ is the cross-entropy loss for classification tasks, $\mathcal{L}_{W}$ is the weight regularization loss defined in Eq. (\ref{eq:w_sum_loss}), and $\mathcal{L}_{A}$ is the pre-activation normalization loss defined in Eq. (\ref{eq:pre_acti_loss}). $\lambda_W$ and $\lambda_A$ are hyperparameters that balance the contribution of each individual loss term. Additionally, the hard constraints are imposed after each weight update to ensure a unity weight sum for all the neurons.

In the second stage, the weights of the pre-trained ANN are directly copied to the SNN to perform inference. We set the threshold of neurons in the last SNN layer to be infinity, such that the decision can be made based on the accumulated membrane potential. 
\textcolor{black}{By directly mapping the pre-activation of ANN into the neuron membrane potential of SNN, it frees the ANN from applying the activation function in the output layer that deteriorates the pre-training.}

\section{Experiments on Image Classification}
\label{exp}
In this section, we first evaluate the effectiveness of the proposed LC-TTFS conversion algorithm on the image classification task. Then, we perform a comprehensive analysis of the conversion efficacy and computational efficiency of converted SNNs. Finally, we present ablation studies that are designed to validate the effectiveness of each individual component of the proposed algorithm. 

\subsection{Experimental Setup}
\label{setup}

\subsubsection{Datasets}
\label{datasets}
We perform image classification experiments on CIFAR-10 and CIFAR-100 \cite{krizhevsky2009learning} datasets, which are commonly used for benchmarking SNN learning algorithms. The CIFAR-10 dataset consists of 60,000 colored images with a standard dataset split of 45,000, 5,000, and 10,000 for train, validation, and testing. These images are categorized into 10 classes with an image size of 32$\times$32$\times$3. The CIFAR-100 dataset is an extended version of CIFAR-10, which includes the 32$\times$32$\times$3 images from 100 classes. We follow the same image pre-processing procedures as outlined in \cite{wu2021progressive}. 

\subsubsection{Implementation details}
\label{implementation}
To facilitate the comparison with other existing works, we adopt VGGNet \cite{simonyan2014very} and ResNet \cite{he2016deep}. In particular, we follow the same VGG-11, VGG-16, and ResNet-20 network architectures as described in \cite{rathi2020enabling}, except that we do not include BN layers. 
The dropout is applied after every layer except the pooling layers in VGG-11 and VGG-16, whereas it is only applied after the fully connected (FC) layers in ResNet-20. Given the absence of BN layers, it is important to have a proper weight initialization. To this end, we initialize the weights of convolutions layers
to the normal distribution $\mathcal{N}(0, \frac{2}{k^2n})$, where $k$ and $n$ correspond to the kernel size and the number of output channels at each layer. For FC layers, the weights are initialized to the normal distribution $\mathcal{N}(0, 0.0001)$.

We use the PyTorch library for all experiments, as it supports accelerated and memory-efficient training on multi-GPU machines. We train VGG-11 and VGG-16 for 300 epochs and ResNet-20 for 200 epochs, and we adopt the SGD optimizer with a momentum of $0.9$ and weight decay of $0.0005$. We initialize the learning rate at $0.01$ and decay its value by 10 at 0.6, 0.8, and 0.9 of the total number of epochs. We follow a discrete-time simulation for SNNs, with a time step of 0.02 ms and 50 time steps for each layer.  We report the classification accuracy on the test set. 

\subsection{Experimental Results and Analysis}
\label{result}
\subsubsection{Classification Accuracy}
Table \ref{tab:result} reports our experimental results on CIFAR-10 and CIFAR-100 datasets. For the CIFAR-10 dataset, our VGG-11 and ResNet-20 models achieved a \textcolor{black}{SNN model} classification accuracy of 91.25\% and 92.67\%, outperforming all other existing conversion methods. Our VGG-16 model also achieved competitive accuracy to other prior works using the same network structure. The same conclusion can also be drawn from the results of the CIFAR-100 dataset. Our ResNet-20 model achieves an accuracy of 72.36\%, which is the best-reported result on this dataset as far as we know. Similarly, our VGG-16 model outperforms other works using temporal coding, which is also competitive to methods based on rate coding.  

\begin{figure}[t]
\centering
\includegraphics[scale=0.35]{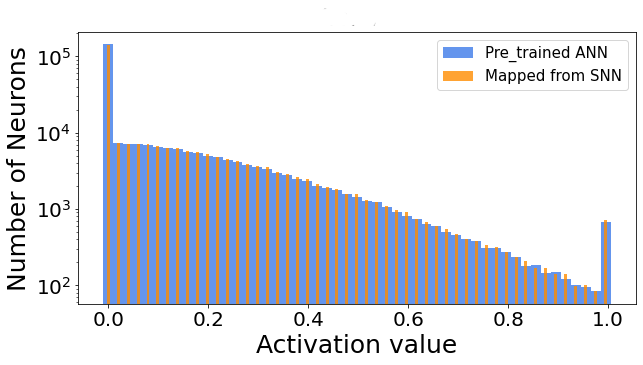}
\caption{Comparison of the distribution of activation values of a pre-trained ANN and those mapped back from the spike times of the converted SNN following the TTFS encoding scheme. The data is extracted from a randomly selected network layer trained on the CIFAR-10 dataset.}
\label{fig: acti_conv}
\end{figure}

The efficacy of our algorithm in achieving a lossless TTFS conversion is pronounced when looking at the conversion error. Notably, as shown in the last column of Table \ref{tab:result}, the conversion errors from the pre-trained ANNs are negligible. To better understand the origin of the conversion errors, we have plotted the activation values of the pre-trained ANN against those mapped back from the spike times of the converted SNN (following the TTFS encoding scheme). As illustrated in Fig. \ref{fig: acti_conv}, the data distribution of the mapped back activation values closely follows that of the pre-trained ANN, except for the quantization errors arising from the discretized SNN simulation. The effect of such quantization errors is marginal from our experimental results and can be easily addressed by increasing the temporal resolution of SNN. 

\begin{table*}[h]
\normalsize
\centering
\caption{Comparison of the classification accuracy of different SNN models on the CIFAR-10 and CIFAR-100 datasets. Note that \textcolor{black}{Acc. refers to the accuracy of converted SNN, and} $\Delta$ Acc. refers to the difference between the pre-trained ANN and the converted SNN.}
\label{tab:result}
\resizebox{16 cm}{!}{
\begin{tabular}{l l l c c c c}
\hline
\hline
\textbf{Dataset}  & \textbf{Method}           &  \textbf{Network Architecture}   & \textbf{Neural Coding} & \textbf{Acc. (\%)} & \textbf{$\Delta$ Acc. (\%)} \\
\hline
\multirow{14}{*}{CIFAR10}   & SPIKE-NORM\cite{sengupta2019going}    & VGG-16         & Rate        & 91.55    & -0.15 \\
   & PTL \cite{wu2021progressive}         & VGG-11         & Rate        & 91.24    & 0.65 \\
   & CQ trained SNN \cite{yan2021near}        & VGG-11        & Rate        & 82.09     & -0.05 \\
   & CQ trained SNN \cite{yan2021near}       & VGG-16        & Rate        & 92.48     & -0.08 \\
   & RMP \cite{han2020rmp}      & VGG-16      & Rate        & 93.63     & -0.01 \\
   & RMP \cite{han2020rmp}      & ResNet-20   & Rate        & 91.36     & -0.11 \\
   & Hybrid Training \cite{rathi2020enabling}     & VGG-16         & Rate        & 91.13    & -1.68 \\
   & Hybrid Training \cite{rathi2020enabling}     & ResNet-20      & Rate        & 92.22    & -0.93 \\
   & T2FSNN \cite{park2020t2fsnn}     & VGG-16          & Temporal    & 91.43    & - \\
   & TSC \cite{han2020deep}      & VGG-16    & Temporal    & 93.63   & -0.01 \\
   & TSC \cite{han2020deep}     & ResNet-20  & Temporal    & 91.42   & -0.05 \\
   & \textbf{Ours}   & \textbf{VGG-11} & \textbf{Temporal}    & \textbf{91.25}     & \textbf{-0.05} \\
   & \textbf{Ours}   & \textbf{VGG-16} & \textbf{Temporal}    & \textbf{92.72}     & \textbf{-0.07} \\
   & \textbf{Ours}   & \textbf{ResNet-20} & \textbf{Temporal}    & \textbf{92.67}  & \textbf{-0.05} \\
\hline
\multirow{9}{*}{CIFAR100}   & CQ trained SNN \cite{yan2021near}      & VGG-like  & Rate        & 71.52     & -0.4 \\
   & RMP \cite{han2020rmp}     & VGG-16    & Rate        & 70.93    & -0.29 \\
   & RMP \cite{han2020rmp}     & ResNet-20   & Rate        & 67.82  & -0.9 \\
   & Hybrid Training \cite{rathi2020enabling}    & VGG-11    & Rate        & 67.87     & -3.34 \\
   & T2FSNN \cite{park2020t2fsnn}     & VGG-16    & Temporal    & 68.79     & - \\
   & TSC \cite{han2020deep}      & VGG-16   & Temporal    & 68.18    & 0.25 \\
   & TSC \cite{han2020deep}     & ResNet-20  & Temporal    & 70.97   & 0.54 \\
   & \textbf{Ours}  & \textbf{VGG-16}   & \textbf{Temporal}    & \textbf{70.15}   & \textbf{-0.13} \\
   & \textbf{Ours}  & \textbf{ResNet-20}  & \textbf{Temporal}    & \textbf{72.36}     & \textbf{0.13} \\
\hline
\hline
\end{tabular}
}
\end{table*}

\subsection{Computational Efficiency}
\label{comp_efficiency}

\textcolor{black}{The TTFS-based SNNs are believed to be computationally more efficient than their rate-based counterparts. To shed light on this point, we follow the practices adopted by Intel \cite{timcheck2023intel} and compute the power proxy via the following equations:
\begin{equation}
	P_{proxy} = SynOPs + NeuronOPs,
\end{equation}
where $SynOPs$ and $NeuronOPs$ are the total number of synaptic operations and the total number of neuron updates, respectively. As extrapolated from the Loihi architecture \cite{loizou2007speech}, the energy weightage of a singular NeuronOPs corresponds approximately to that of around 10 SynOPs. This is due to the NeuronOPs being a multi-bit operation. 
Importantly, our utilized neuron model, ReL-PSP, avoids the exponential leaky mechanism present in the LIF model. 
To update the neuronal state, ReL-PSP requires only an addition operation at each time step, which is almost equivalent to a SynOPs.} 

\textcolor{black}{
With reference to the configuration specifics reported by Lee et al. \cite{lee2020enabling}, we have computed and tabulated the total $P_{proxy}$ cross our work and that of Diehl et al. \cite{diehl2015fast}, and Segupta et al. \cite{sengupta2019going}. As illustrated in Table \ref{tab:P_proxy}, our TTFS conversion algorithm evidently surpasses the energy efficiency metrics of traditional rate-based conversion methods.}

\begin{table}[h]
\color{black}
\normalsize
\centering
\caption{Comparison of the Total $P_{proxy}$.}
\label{tab:P_proxy}
\begin{tabular}{l c c c}
\hline
\hline
Model       & Diehl et al.     & Segupta et al.      & Ours \\ 
\hline
VGG9       & 730.4M   & 794M         & 117M\\
ResNet9    & 956.4M      & 1,226.6M     & 107.8M\\
ResNet11   & 854.6M      & 796.3M       & 144.4M\\
\hline
\hline
\end{tabular}
\end{table}

\subsection{Ablation Studies}
\label{ablation}
Here, we present the ablation studies that are designed to validate the necessity and contribution of each individual component in the proposed algorithm. All the experiments are performed on the CIFAR-10 dataset using VGG-11. The experimental results are summarized in Table \ref{tab:ablation} with more detailed discussions presented in the following. 
\begin{table*}[h]
\normalsize
\centering
\caption{Summary of the result of ablation studies. Note that $\Delta$ Acc. refers to the difference between the pre-trained ANNs and the converted SNNs given in columns 7 and 8, respectively. \textbf{Soft}: soft constraint for weight sum; \textbf{Hard}: hard constraint for weight sum; \textbf{ReLU1}: ReLU1 activation function; \textbf{Dynamic}: dynamic firing threshold; \textbf{Norm}: pre-activation normalization.}

\label{tab:ablation}
\begin{tabular}{c c c c c c c c c}
\hline
\hline
\textbf{Model}   & \textbf{Soft} & \textbf{Hard}  & \textbf{ReLU1}  & \textbf{Norm} & \textbf{Dynamic}        & \textbf{ANN (Acc.\%)}   & \textbf{SNN (Acc.\%)}     & \textbf{$\Delta$ Acc.\%} \\ 
\hline
Baseline       & \Checkmark & \Checkmark & \Checkmark & \Checkmark & \Checkmark     & 91.30            & 91.25        & 0.05\\
1       & \XSolidBrush & \XSolidBrush & \XSolidBrush & \XSolidBrush & \XSolidBrush           & 91.33            & 10.00        & 81.33\\
2a       & \Checkmark & \XSolidBrush & \Checkmark & \Checkmark & \Checkmark         & 91.57            & 86.95        & 4.62\\
2b       & \XSolidBrush & \XSolidBrush & \Checkmark & \Checkmark & \Checkmark         & 92.10            & 29.18        & 62.92\\
3       & \Checkmark & \Checkmark & \XSolidBrush & \Checkmark & \Checkmark             & 90.87           & 64.44        & 26.43\\
4       & \Checkmark & \Checkmark & \Checkmark & \XSolidBrush & \Checkmark          & 89.60            & 89.67        & -0.07\\
5       & \Checkmark & \Checkmark & \Checkmark & \Checkmark & \XSolidBrush           & 91.30            & 77.82        & 13.48\\
\hline
\hline
\end{tabular}
\end{table*}

\subsubsection{Model 1}
\label{ab_constraints}
We began by removing all the constraints that applied to the pre-trained ANN (i.e., soft and hard constraints for the weight sum, ReLU1 activation function, and pre-activation normalization) and the converted SNN (i.e., dynamic firing threshold). As a result, the accuracy of the ANN improved by only 0.03\% over the baseline model. It suggests applying the proposed set of constraints has negligible influence on the ANN pre-training. The converted SNN, however, failed to perform the image classification task with the test accuracy dropping to a chance level, indicating it is crucial to apply the proposed set of constraints for a lossless TTFS conversion.

\subsubsection{Model 2a and 2b}
\label{ab_soft}
We further studied the necessity of soft and hard constraints for the weight sum regularization. Dropping the hard constraint while keeping all the rest constraints during the ANN pre-training had minimal impact on the ANN model, whereas the accuracy of the converted SNN dropped by 4.62\%. It suggests the hard constraint for weight sum regularization is essential to eliminate the time-warping problem discussed in Section \ref{hardCons}. We further explored dropping both the soft and hard constraints during the ANN pre-training. Interestingly, this allows the ANN model to perform better than the baseline model. However, we noticed that the accuracy of the converted SNN dropped by 62.92\%, implying the soft constraint is critical for alleviating the time-warping problem. 

\subsubsection{Model 3}
\label{ab_relu}
ReLU1 activation function bounds the activation values within a particular interval, so as to establish a one-to-one correspondence between the ANN and SNN. To understand the contribution of this constraint, we replaced the ReLU1 activation function with a standard ReLU function, while keeping all the rest constraints during the ANN pre-training. Interestingly, compared to the baseline model with full constraints, the model using the ReLU function performs slightly worse than the one using the ReLU1 function. This may be explained by the fact that the ReLU1 activation function can indirectly regularize the weight sum and hence lead to easier fulfillment of the weight sum constraint. However, when changed to the ReLU activation function, it results in a severe accuracy drop of 26.43\% on the converted SNN. These results indicate the ReLU1 function is imperative to ensure a one-to-one correspondence between the activation value of ANNs and the spike time of SNNs.

\subsubsection{Model 4}
\label{ab_pre-acti}
Similar to Models 2 and 3, we pre-trained the ANN  with all constraints except for the pre-activation normalization. This relaxes the pre-activation values from falling into the value range desired by the ReLU1. It, therefore, leads to poorer accuracy in the pre-trained ANN. We notice that the accuracy drop is more significant for networks with more layers (data not shown here), which is due to the internal covariate shift problem discussed in Section \ref{pre_acti}. As expected, the absence of the pre-activation normalization technique did not affect the SNN conversion, and it can still achieve comparable accuracy to the pre-trained ANN.

\subsubsection{Model 5}
\label{ab_dynamic}
\begin{figure}[t]
    \centering  
	\subfloat{
	\includegraphics[scale=0.25]{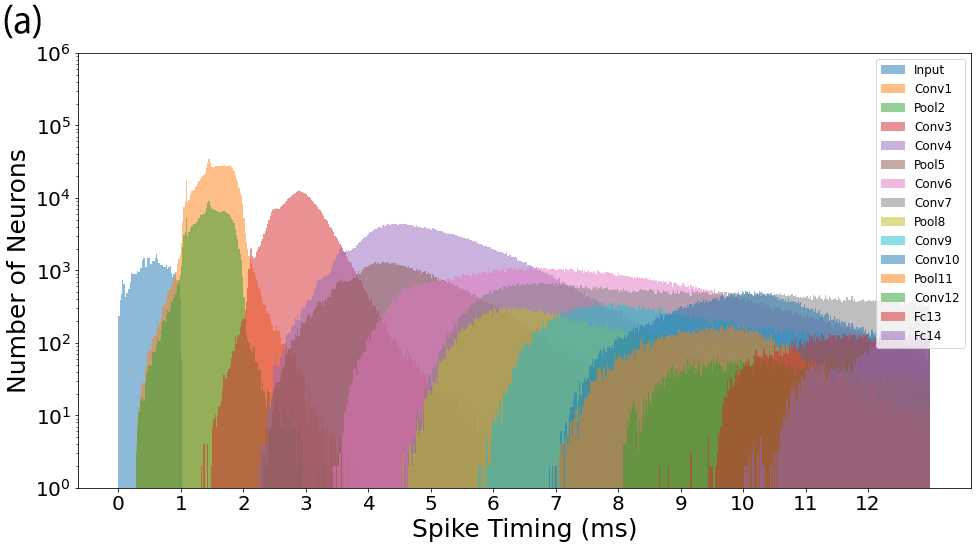}}\hspace{0em}
	
	\subfloat{
	\includegraphics[scale=0.25]{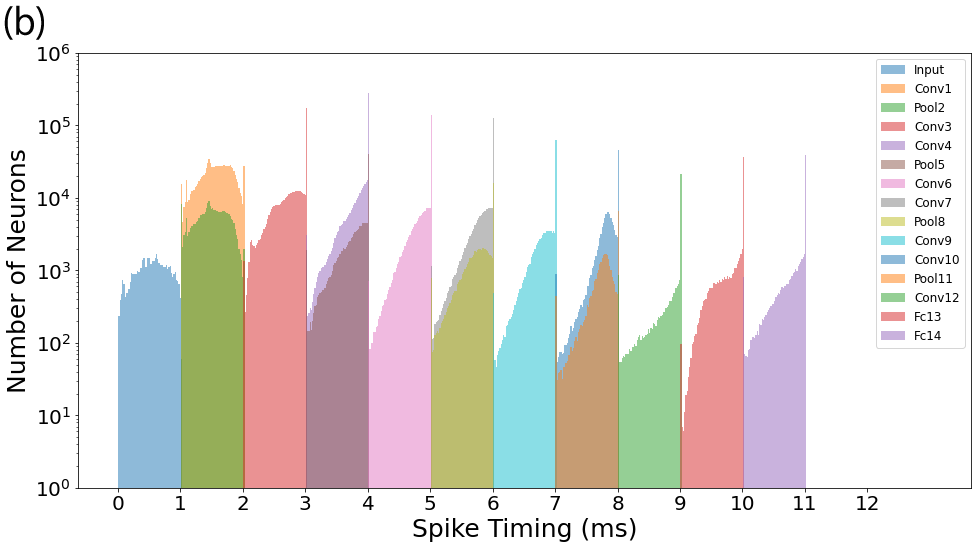}}\hspace{0em}
    \caption{Illustration of spike timing distribution of SNN models (a) with a fixed firing threshold and (b) with a dynamic firing threshold.}
    \label{fig: timing_dist}
\end{figure}

Finally, we follow the same ANN pre-training procedures as the baseline model, while during the network conversion, we replaced the dynamic firing threshold with a fixed threshold of $1$. This modification results in a 13.48\% accuracy drop during the network conversion. To further illustrate the effectiveness of applying the dynamic threshold to address the temporal dynamics problem discussed in Section \ref{premature}, we plotted the spike time distribution using the fixed threshold and the proposed dynamic threshold. As shown in Fig. \ref{fig: timing_dist} (a), the spike times spread outside their allocated time intervals, and this problem becomes more obvious for deeper layers. Consequently, it leads to a high level of mismatch from the pre-trained ANN. In contrast, with the proposed dynamic threshold, the neurons at different layers only spike in their allocated time interval, i.e., $[Tl, T(l+1))$. This ensures the pre-synaptic spiking neurons fully contribute to their post-synaptic spiking neurons, eliminating the temporal dynamics problem. \textcolor{black}{Although this non-overlapping time window may scarify the synchronized benefit of SNN, we highlight that our goal is to achieve lossless conversion from ANN to SNN using temporal coding without any training in SNN. To this end, the ANN activation and SNN spiking time are needed to be perfectly matched with each other. Therefore, we adopt the proposed synchronized approach to realize this requirement. It is worth noting that the synchronized layer-wise processing is also used in other exciting temporal coding works, such as \cite{zhang2019tdsnn, park2020t2fsnn, park2021training}.}

\section{Experiments on Signal Reconstruction}
\label{regression}
In the previous section, we demonstrate the effectiveness and scalability of the proposed LC-TTFS conversion algorithm on image classification tasks. 
\textcolor{black}{Existing TTFS-based learning algorithms often employ methods 
that achieve TTFS-based learning by either dropping or masking subsequent spikes after the first one during the training phase.
Moreover, while there are gradient-based direct training algorithms cited in [45, 48], they have manifested convergence challenges, especially concerning deep neural networks. Although these methods might prove efficient for classification tasks, their efficacy diminishes notably when deployed for signal reconstruction tasks.
In contrast, our proposed method adopts a conversion-based approach. This approach fundamentally eradicates the convergence issue and paves the way for establishing a direct and lossless mapping from the ANN. This ensures that information integrity remains uncompromised throughout the process.
}
Performing TTFS encoding at each layer, allows us to preserve the information across the network and opens up the opportunity to perform signal reconstruction tasks with TTFS-based SNNs.
In this section, we demonstrate the applicability and superior learning capability of the proposed LC-TTFS algorithm on two signal reconstruction tasks: image reconstruction and speech enhancement tasks.

\subsection{Image Reconstruction with Autoencoder}
\label{reg_AE}
Here, we first demonstrate the applicability of our algorithm to image reconstruction tasks with an autoencoder network. The autoencoder is a typical neural network that is used to learn compact latent representations of input signals via a bottleneck layer that has a reduced latent dimension. From this compact latent representation, the autoencoder then reconstructs the original input signals as accurately as possible \cite{goodfellow2016deep}.

\subsubsection{Experimental Setup}
\label{AE_exp}
\textcolor{black}{The experiments on image reconstruction are performed using the MNIST dataset \cite{lecun1998gradient}. Following the approach from \cite{wu2021progressive}, we use a fully-connected autoencoder with a 784-128-64-32-64-128-784 architecture and train it to minimize the MSE between the original input and the reconstructed output signal.}

\textcolor{black}{The initial model was pre-trained as an ANN before being converted into a TTFS-based SNN using our proposed algorithm. We utilized the SGD optimizer for pre-training, with a cosine annealing learning rate schedule \cite{loshchilov2016sgdr}. The output from spiking neurons was processed through a sigmoid function to generate the reconstructed image.}

\textcolor{black}{Evaluation of our model was carried out using two commonly used image quality metrics, PSNR and SSIM, both of which provided insight into the quality of reconstructed images. We provide the results of these evaluations on the test set.}

\subsubsection{Result and Analysis}
\label{AE_results}
\begin{figure}[t]
\centering
\includegraphics[scale=0.7]{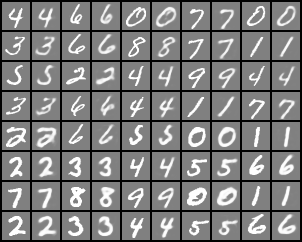}
\caption{Comparison of the original and the reconstructed images from our TTFS-based autoencoder. For each image pair, the left one is the original image, and the right one is the reconstructed image.}
\label{fig: AE_image}
\end{figure}

Table \ref{tab:AE_result} summarises the results of on the image reconstruction task. Our TTFS-based SNN model achieves a comparable performance to the ANN-based counterpart in terms of the PSRN and SSIM metrics. In addition, the qualitative results shown in Fig. \ref{fig: AE_image} demonstrate the TTFS-based SNN model can effectively reconstruct the images with high quality. Altogether, these results suggest the proposed conversion algorithm is highly effective for the image reconstruction task. 

By representing information using spike times, the TTFS-based SNN is expected to greatly improve the energy efficiency over the rate-based counterparts. 
\textcolor{black}{Following the same evaluation metrics introduced in Section \ref{comp_efficiency}, we report the energy proxy in Table \ref{tab:AE_result}. As shown, the energy efficiency of our TTFS-based model remains competitive with a high-optimized rate-based SNN reported in \cite{wu2021progressive}.}

\begin{table}[htb]
\normalsize
\centering
\caption{Comparison of the results of different methods on the image reconstruction task.}
\label{tab:AE_result}
\begin{tabular}{l l c c c}
\hline
\hline
\textbf{Model}               &  \textbf{Coding}    & \textbf{PSNR}      & \textbf{SSIM}  & \textbf{\textcolor{black}{Energy Proxy}}\\ 
\hline
SNN \cite{wu2021progressive}    & Rate          & 20.74     & 0.84      & \textcolor{black}{715.8K}\\
\textbf{ANN (ours)}             & -             & 20.90     & 0.917     & \textcolor{black}{-} \\
\textbf{SNN (ours)}             & Temporal      & 20.83     & 0.916     & \textcolor{black}{581.2K}\\
\hline
\hline
\end{tabular}
\end{table}

\subsection{Time-domain Speech Enhancement}
\label{reg_tasks}
Speech enhancement, which improves the intelligibility and quality of noisy speech signals \cite{loizou2007speech}, has proven its importance in a vast amount of applications, including voice-based communication, hearing aids, speech recognition, and speaker identification \cite{yang2005spectral, yu2008minimum, maas2012recurrent, ortega1996overview, 9502020}. Conventional speech enhancement methods are typically based on statistical models that estimate the data distribution of clean and noise signals, such as spectral subtraction \cite{berouti1979enhancement} and Wiener filter \cite{lim1978all}. However, these methods have shown limited improvements in speech quality under real-world scenarios. Recently, the ANN-based speech enhancement models have greatly improved the speech quality and intelligibility under complex acoustic environments \cite{pascual2017segan, shah2018time, choi2018phase, kim2020t, abdulbaqi2019rhr}. Given a huge demand for speech enhancement technologies on mobile and IoT devices that have a limited power budget, it is, therefore, beneficial to develop power-efficient SNN-based speech enhancement models. 

\begin{figure*}[htb]
    \centering  
	\subfloat[]{
	\includegraphics[scale=0.52]{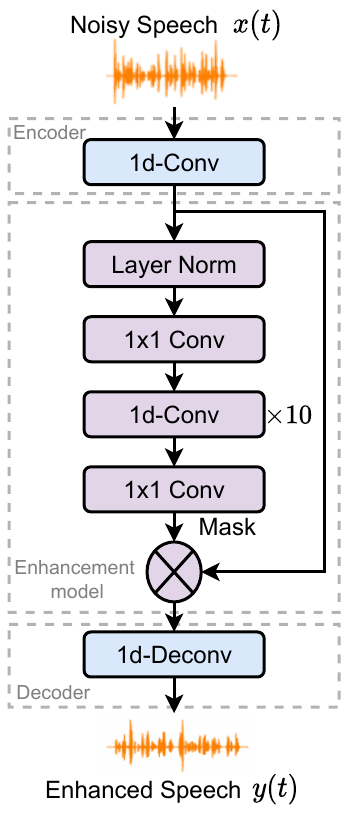}}\hspace{3em}
	\subfloat[]{
	\includegraphics[scale=0.215]{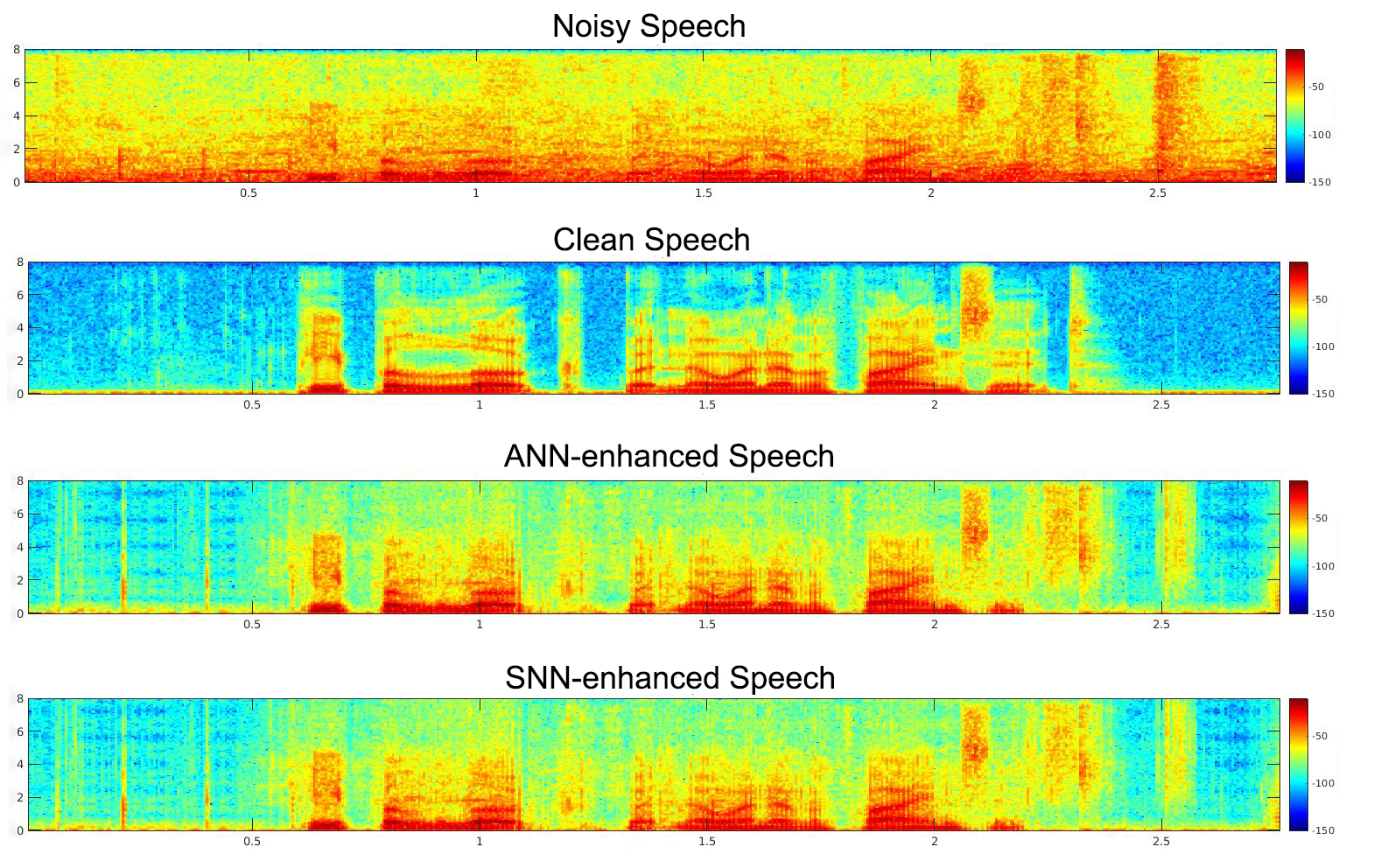}}\hspace{0em}
    \caption{(a) SNN-based speech enhancement network architecture. (b) From top to bottom: the power spectrum of noisy, clean, ANN-enhanced, and SNN-enhanced speech waveforms.} 
    \label{fig:speech_enhancement}
\end{figure*}

Speech enhancement can be considered as separating the human voice from the background noise. Inspired by the recent success of the time-domain speech separation model ConvTasNet \cite{luo2019conv}, we proposed a TTFS-based speech enhancement model. As illustrated in Fig. \ref{fig:speech_enhancement}(a), the proposed speech enhancement model takes the noisy speech waveform as input and outputs an enhanced speech waveform. This model consists of three parts: an encoder, an enhancement module, and a decoder. The \textbf{encoder} transforms the noisy speech waveform $x(t)$ into a high-dimensional feature representation, namely embedding coefficients, using a 1D convolutional layer. The 1D convolutional layer contains $N(=128)$ filters, and each filter is configured to have a time window of $L(=20)$ and a stride of $L/2(=10)$ samples. For each time window, the
\textbf{enhancement module} estimates a mask, using a stack of dilated convolutional layers, to separate the noise from the human voice. A $1 \times 1$ convolution layer is first applied to normalize the encoded feature representation, and the dilated convolution layers with the filter of 128, kernel size of $1 \times 3$, and stride of $1$ are repeated $10$ times with doubled dilation rate of $[1, 2, 4, ..., 512]$. The mask for human voices is then estimated by applying another $1 \times 1$ convolution layer. Subsequently, the feature representation of the enhanced speech is obtained by masking the background noise with the estimated mask.
Finally, the \textbf{decoder} reconstructs a high-quality speech waveform $y(t)$ from the masked feature representation using a 1D deconvolutional layer, which takes a reverse operation to the 1D convolutional layer in the encoder. 

Following the ConvTasNet, we train the proposed SNN-based speech enhancement model to minimize the multi-scale scale-invariant signal-to-distortion ratio (SI-SDR)\cite{le2019sdr} loss, which is defined as: 
\begin{equation}
    \mathcal{L}_{SI-SDR} = - 10log_{10} \biggl( \frac{\Vert {\frac{\langle \hat{s}, s\rangle}{\langle s, s\rangle} s}\Vert^2}{\Vert {\frac{\langle \hat{s}, s\rangle}{\langle s, s\rangle} s - \hat{s}}\Vert^2} \biggl)
	\label{eq: sisdr}
\end{equation}
\noindent where $\hat{s}$ and $s$ are enhanced and reference clean speech signals, respectively. We normalize these two signals to zero means to ensure scale invariance. 

\subsubsection{Experimental Setup}
\label{reg_setup}
\textcolor{black}{To test our speech enhancement model, we employed a widely recognized dataset by Valentini et al. \cite{valentini2016investigating}. This dataset includes clean utterances from the Voice Bank corpus \cite{veaux2013voice} and its noisy version created by combining clean utterances with environmental noise samples. 
The training set comprises 11,572 utterances mixed with ten types of noise at four different SNRs: 15 dB, 10 dB, 5 dB, and 0 dB. The test set contains 824 distinct utterances blended with five additional noise types at four SNRs: 17.5 dB, 12.5 dB, 7.5 dB, and 2.5 dB. Following the precedent set by SEGAN \cite{pascual2017segan} and CNN-GAN \cite{shah2018time}, we reduced the original sampling rate to 16 kHz, without additional pre-processing.}

\textcolor{black}{Using the LC-TTFS algorithm, we pre-trained the ANN-based speech enhancement model for 100 epochs, utilizing an early stopping scheme and an Adam optimizer. The model was then converted into an SNN-based module, employing the membrane potential in the $1 \times 1$ spiking convolution layers.}

We evaluate the speech enhancement models using the following standard metrics, which are available on the publisher's website\footnote{\url{https://www.crcpress.com/downloads/K14513/K14513_CD_Files.zip}}.
\begin{enumerate}
    \item PESQ: Perceptual evaluation of speech quality. The wide-band version recommended by ITU-T P.862.2 standard \cite{rec2005p} is used in this work.
    \item CSIG: Mean option score (MOS) prediction of the signal distortion attending only to the speech signal \cite{hu2007evaluation}.
    \item CBAK: MOS prediction of the intrusiveness of background noise \cite{hu2007evaluation}.
    \item COVL: MOS prediction of the overall effect \cite{hu2007evaluation}.
\end{enumerate}
All metrics are calculated by comparing the enhanced speech to the clean reference speech, we report the average values for all 824 utterances in the test set.

\subsubsection{Result and Analysis}
\label{reg_result}
Table \ref{tab:SE_result} compares the results of our ANN- and SNN-based speech enhancement models with other existing works using the four evaluation metrics introduced earlier. Our ANN-based model outperforms other existing methods across all the evaluation metrics, suggesting the effectiveness of the proposed model architecture. Moreover, the TTFS-based SNN model achieved comparable performance to the pre-trained ANN model, demonstrating the capability of the proposed TTFS conversion algorithm in solving the challenging speech enhancement task.  

We also performed subjective evaluation by listening to the enhanced speech signals generated by both the ANN- and SNN-based speech enhancement models. We find the SNN-enhanced speech samples are nearly indistinguishable from those high-quality ones generated from the ANN model. We publish some enhanced speech examples from the test set online to demonstrate our model performance\footnote{The listening examples are available online at \url{https://drive.google.com/file/d/1g5OCAtYH1B3tJE5_zGqDZam8bMcnIEvR/view?usp=sharing}}. In addition, we select a random speech sample from the test set and plot the power spectrum for the corresponding noisy, clean, ANN- and SNN-enhanced speech waveforms as shown in Fig. \ref{fig:speech_enhancement}(b). It is clear that the SNN-enhanced speech spectrum exhibits a high level of similarity to the ANN-enhanced one, and both of them are very close to the ground truth clean speech spectrum. These results again highlight the effectiveness of the proposed model architecture and the TTFS conversion algorithm.

\begin{table}[h]
\normalsize
\centering
\caption{Comparison of the experimental results of different speech enhancement models. \textcolor{black}{The results are higher the better.}}
\label{tab:SE_result}
\begin{tabular}{l c c c c}
\hline
\hline
Model       & PESQ      & CSIG      & CBAK      & COVL \\ 
\hline
Noisy       & 1.97      & 3.35     & 2.44       & 2.63\\
Wiener \cite{pascual2017segan}      & 2.22      & 3.23     & 2.68       & 2.67\\
SEGAN \cite{pascual2017segan}       & 2.16      & 3.48     & 2.94       & 2.80\\
CNN-GAN \cite{shah2018time}     & 2.34      & 3.55     & 2.95       & 2.92\\
\textbf{ANN (Ours)}  & \textbf{2.36}      & \textbf{3.63}     & \textbf{3.03}       & \textbf{2.98}\\
\textbf{SNN (Ours)}  & \textbf{2.35}      & \textbf{3.64}     & \textbf{3.01}       & \textbf{2.98}\\
\hline
\hline
\end{tabular}
\end{table}

\section{Discussion and Conclusion}
\label{conclusion}
In this work, we identify and thoroughly investigate \textcolor{black}{two} major problems underlying the effective TTFS conversion, namely the \textcolor{black}{temporal dynamics problem and time-warping problem}. Based on this study, we further propose a novel TTFS conversion algorithm to address these problems, namely LC-TTFS. Firstly, to tackle the temporal dynamics problem, we introduce a dynamic firing threshold mechanism for spiking neurons that only allows neurons to fire within the allocated time window. In this way, the causal relationship between the input and output neurons is maintained throughout the network layers. Secondly, we apply a set of well-designed loss functions during the ANN pre-training to eliminate the time-warping problem. Finally, we apply the pre-activation normalization technique during the ANN pre-training to alleviate the internal covariate shift problem \textcolor{black}{due to lack of BN layer}. With these problems being well addressed, we establish a near-perfect mapping, apart from the marginal discretization error in the SNN simulation, between the ANN activation values and the SNN spike times, leading to a near-lossless TTFS conversion. This also enables us to go beyond the commonly considered classification tasks and opens up a new avenue for solving high-fidelity signal reconstruction tasks with TTFS-based SNNs. 

The SNNs thus converted have demonstrated superior classification and signal reconstruction capabilities on image classification, image reconstruction, and challenging speech enhancement tasks. By representing information using spike times, instead of firing rates, we show TTFS-based SNNs can significantly improve the computational efficiency over both ANN and rate-based SNNs. By avoiding costly and ineffective direct SNN training, the proposed algorithm, therefore, opens up myriad opportunities for deploying efficient TTFS-based SNNs on power-constrained edge computing platforms. The carefully designed ablation studies on each individual component of the proposed algorithm highlight the necessity and synergy of these algorithmic components in achieving a near-lossless TTFS conversion. We would like to acknowledge that the proposed TTFS-based SNN requires a separate and non-overlapping time window for each layer, which adversely affects the inference speed. Therefore, we are interested in studying whether a perfect mapping can still be achieved under a shared time window to improve the inference speed, and we leave this as future work.




 
%
\bibliographystyle{IEEEtran} 
\bibliography{citation}

\vfill

\end{document}